\documentclass[sigconf]{acmart}

\usepackage{booktabs} 
\usepackage{algorithm2e} 
\usepackage{gensymb}

\setcopyright{rightsretained}

\acmDOI{10.475/123_4}

\acmISBN{123-4567-24-567/08/06}

\acmConference[ETRA 2018]{ACM Symposium on Eye Tracking Research \& Applications}{June 2018}{Warsaw, Poland} 
\acmYear{2018}
\copyrightyear{2018}

\acmArticle{4}
\acmPrice{15.00}

\begin{document}
\title{Tertiary Eye Movement Classification by a Hybrid Algorithm }

\author{Samuel-Hunter Berndt}
\affiliation{
  \institution{Michigan State University}
  \city{East Lansing} 
  \state{Michigan} 
  \postcode{48824}
}
\email{berndts2@msu.edu}

\author{Douglas Kirkpatrick}
\affiliation{
  \institution{Michigan State University}
  \city{East Lansing} 
  \state{Michigan} 
  \postcode{48824}
}
\email{kirkpa48@msu.edu}

\author{Timothy Taviano}
\affiliation{
  \institution{Michigan State University}
  \city{East Lansing} 
  \state{Michigan} 
  \postcode{48824}
  }
\email{tavianot@msu.edu}

\author{Oleg Komogortsev}
\affiliation{
  \institution{Michigan State University}
  \city{East Lansing} 
  \state{Michigan} 
  \postcode{48824}
  }
\email{ok@msu.edu}

\begin{abstract}
The proper classification of major eye movements, saccades, fixations, and smooth pursuits, remains essential to utilizing eye-tracking data. There is difficulty in separating out smooth pursuits from the other behavior types, particularly from fixations. To this end, we propose a new offline algorithm, I-VDT-HMM, for tertiary classification of eye movements. The algorithm combines the simplicity of two foundational algorithms, I-VT and I-DT, as has been implemented in I-VDT, with the statistical predictive power of the Viterbi algorithm. We evaluate the fitness across a dataset of eight eye movement records at eight sampling rates gathered from previous research, with a comparison to the current state-of-the-art using the proposed quantitative and qualitative behavioral scores. The proposed algorithm achieves promising results in clean high sampling frequency data and with slight modifications could show similar results with lower quality data. Though, the statistical aspect of the algorithm comes at a cost of classification time.
\end{abstract}

%
%



\maketitle

\section{Introduction}
In order to further advance eye tracking research and push forward the use of eye trackers in industry, the identification of eye movements is highly desired. To date there are six primary types of eye movements exhibited by the human oculo-motor system (HOS): fixations, saccades, smooth pursuits (SPs), optokinetic reflex, vestibulo-ocular reflex, as well as vergence \cite{leigh2015neurology}. Of these, fixations, saccades, and SPs are the most frequently studied. Intuitively, fixations are what one experiences when staring at a stationary object. Due to such, it can be easily classified using an overall relative velocity of zero degrees per second, or alternatively using a net zero movement over a period of time. As noise is likely to be present in most eye movement samples, algorithms that employ single threshold classification, especially if the assumption is made that fixations will have zero velocity, may require adjustment on a sample by sample basis. Saccades are where one quickly moves their eyes from one object to another, commonly exhibiting speeds in excess of 300 degrees per second  \cite{dodge1901angle}. Thus, classifying saccades based off of a high overall velocity between points can be an effective method. SPs are a function of the HOS that can be described as when ones eyes attempt to maintain high acuity on a moving object \cite{duchowskit, poole2005lj}. Occasionally one would get distracted or lose focus of the object, in which case a catch up saccade may be exhibited to regain focus. As SPs have variability in their speed, identifying them while also in the presence of fixations can be difficult. This is especially pertinent when noise is present in the data. As fixations are exhibited when someone is focused on an object, they are frequently used in human-computer interaction applications as a selection method \cite{istance2010designing}. Additionally researchers have found abnormalities in saccade and SPs that have led to the diagnosis of some pathologies of the HOS \cite{isotalo2009oculomotor}. Thus the continued development of methods for the classification of these eye movement types is still thought to be an important area of research.

I-VT and I-DT are foundational threshold algorithms used to separate fixations from saccades. I-VT uses a velocity threshold that takes advantage of the large distinction in velocity between the fast moving saccade and the relatively stationary fixation \cite{Salvucci2000}. I-DT uses two thresholds to make the distinction: duration and dispersion. The duration threshold can be dependent on the users and the stimulus. However, the threshold is commonly used with a minimum limit due to the HOS's inability to pick up information in less than 100 ms \cite{widdel1984operational}. Dispersion is the sensitivity of the algorithm in regards to an eye movement's position graphically. It acts as a noise filter by allowing any points within the window to be considered a fixation. I-DT places a window the size of the duration threshold over a series of eye movement points. In the case of a saccade, the points would break the dispersion threshold over that window and be classified as such. Under ideal conditions, the dispersion threshold for a fixation would be zero. However, as noise is inherent in eye movement classification, assuming the threshold is zero can lead to poor results. Classifying saccades from fixations is effective when using approaches that search for single features such as I-VT and I-DT. However, classifying the two from smooth pursuits is not possible using these methods alone. Two algorithms, I-VVT and I-VDT \cite{Komogortsev2013}, build off of the previous foundational algorithms and incorporate an additional threshold to make this classification. Both I-VVT and I-VDT use an initial velocity threshold to classify between saccades and both fixations and SPs. The remaining points are then classified either using another velocity threshold in I-VVT or using a dispersion threshold as in I-VDT. 

Threshold algorithms have the ability to provide promising results. However, the thresholds used in the algorithms tend to vary based on the input data. In some cases eye trackers provide substantially more noise than others, in which case dispersion and velocity thresholds should be updated accordingly. Thresholds aren't the only method in eye movement classification, however. Other methods previously used in the field include classification using the main-sequence relationship, amplitude-duration relationship, saccades' wave form \cite{leigh2015neurology}, or statistical probabilities \cite{Santini2016}. 

Hidden Markov Models (HMM) are statistical models that attempt to discover hidden states. The forward algorithm, also known as the Viterbi algorithm \cite{forney1973viterbi}, is used to determine the most likely sequence of hidden or unobserved states. This is done by calculating three probabilities: the observational probability of each of \textit{n} states, the transitional probabilities between each state, and the probability at each time step that any of the previous states will lead to the current state. At the beginning of the algorithm two matrices are created: a traceback matrix and an emission matrix. The emission matrix is populated at each step by determining the maximum probability given some transition. The traceback matrix is populated by determining which of the \textit{n} state probabilities is higher at a given time step. Following the completion of the algorithm, an iterator will travel from the last column index in the traceback matrix, starting in the row with the highest probability, and go backward through the traceback matrix using the value of that cell. Effectively, the Viterbi algorithm works by choosing the series of classifications which has the highest probability \cite{forney1973viterbi}. An approach to automatically classify binary eye movements using a HMM has been noted \cite{Salvucci2000} and has been appropriately named I-HMM.

Behavioral scores provide an automated method of creating meaningful classifications when using a step-ramp stimulus \cite{komogortsev2010standardization}. The scores are created under the assumption that the saccade and fixation stimulus are encoded to follow a normal HOS's behavior. It follows that the selected thresholds will hold the same performance when given different stimulus. As the equipment is unlikely to change following the calibration procedure, behavioral scores allow the automated selection of thresholds for classification immediately after calibration. 

As eye tracking is a quickly growing field, meaningful improvements in the classification of eye movements continues to be a sought after goal. In this paper we propose a new hybrid algorithm, I-VDT-HMM, which builds off of the previous work of I-VDT \cite{Komogortsev2013} and I-HMM \cite{Salvucci2000} in an attempt to take the advantages of a threshold algorithm while statistically enhancing our results using the Viterbi algorithm \cite{forney1973viterbi}. To determine the benefits of our algorithm in a variety of conditions, we subsampled the high quality eye tracking data into 8 subsample frequencies and tested it across 8 subjects. Our results are compared against the state of the art I-BDT algorithm \cite{Santini2016}, and the I-VDT algorithm \cite{Komogortsev2013}. Our \textit{MATLAB} implementation of I-VDT-HMM can be found on GitHub at \href{https://github.com/BerndtSam/I-VDT-HMM}{https://github.com/BerndtSam/I-VDT-HMM}. 

\section{Related Work and Evaluated Algorithms}

\subsection{I-VDT}
I-VDT is a seminal algorithm in tertiary eye movement classification \cite{Komogortsev2013}.  The algorithm is a combination of the I-VT and I-DT algorithms, using a velocity threshold to identify saccades, while a moving dispersion window is used to separate fixations and smooth pursuits.  I-VDT is noted for fast evaluation time, accurate classification, as well as ease of implementation \cite{Komogortsev2013}. 

\subsection{I-BDT}
Bayesian Decision Theory Identification, I-BDT, is a probability based algorithm designed for low resolution eye trackers proposed in \cite{Santini2016}. I-BDT requires no calibration as it is based entirely on eye positional data and so is operational at run time. I-BDT uses a bayesian decision theory approach where it relies on a prior and likelihoods to calculate the posterior probability of a classification using velocity and a movement ratio over a temporal window as classification features. I-BDT uses the assumption that the velocity and movement ratio of a fixation must be zero, which will be discussed later. A more robust description of the algorithm can be found in 
\cite{Santini2016}.

\subsection{I-VDT-HMM}
I-VDT-HMM is an offline hybrid algorithm proposed in this work. It is derived from I-VDT \cite{Komogortsev2013} in that it uses a velocity threshold to separate out saccades, and a dispersion window to separate fixations from smooth pursuits. In order to statistically ensure the resulting scores, two two-state Viterbi algorithms are employed; once after I-VT and the other after I-DT. The first HMM, like I-VT, is used to separate saccades from a combination of both fixations and smooth pursuits. In this HMM iteration, the velocity of each state is used as the feature classifier. This works well as both fixations and SPs have relatively the same velocity when compared in a noisy environment, and saccades tend to be much faster than the two. The second HMM is ran after the dispersion window separates fixations from smooth pursuits. This HMM once again has two states; one for fixations and the other for smooth pursuits. In this HMM iteration, similar to I-DT, the dispersion of each class is employed as the feature inputs to our classifier. Thus rather than using the velocity to make classifications, we create a dispersion window for each previously classified fixation and smooth pursuit eye record and then take the difference, the dispersion, between the maximum and minimum. We then take the average and standard deviation of the dispersion as our feature inputs to the PDF function. Upon completion of each HMM, we determine whether a set of epsilon values, the difference between each iteration of a set of variables, has been met. If the change in values are less than the epsilon value, ensuring that the algorithm has converged, it then proceeds onto the next step. As I-VDT-HMM employs the Viterbi algorithm \cite{forney1973viterbi} which requires the iteration over all states, the algorithm is an off-line classification algorithm. The pseudocode for I-VDT-HMM can be found in Algorithms \ref{alg:I-VDT-HMM}, and \ref{alg:viterbi}.

HMMs use three probabilities to determine hidden states, the observed probability, transitional probabilities between states, and the probability that any of the previous states will lead to the current state: the emission probability. The transitional probability is calculated using the sum of the transitions from one particular state to another over the total number of eye records. The calculation for transitional probabilities can be found in Equation \ref{eq:transition}, where \textit{s} is the current state, \textit{p} is the previous state, \textit{n} is the number of eye records, \textit{i} is the index of the iterator iterating the eye records, \textit{$p_0$} is the "from" portion of the transition we're calculating and \textit{$s_0$} is the "to" portion. To calculate the observational probabilities, the probability density function (PDF) using the means and standard deviations of the respective classes are used where the results are then normalized. The equation for the PDF can be found in Equation \ref{eq:pdf} where \textit{$\sigma^2$} is the standard deviation, \textit{$\mu$} is the mean, and \textit{$x$} is the current observation. The emission probability for each state is calculated using the product of the previous emission probabilities, the transitional probabilities given the current and previous states, and the observational probability given the current state. The calculation for the emission probability can be found in Equation \ref{eq:emission}, where \textit{s} is the current state, \textit{p} is the previous state, and \textit{i} is the current state's index.

\begin{equation}
P_{transition, s, p, s_0, p_0} = \frac{1}{n}\sum_{i=1}^{n} X; \\
X = 
    \begin{cases}
      1, & \text{if $p \Rightarrow s \cap p=p_0 \cap s=s_0$} \\
      0, & \text{otherwise}
    \end{cases}
\label{eq:transition}
\end{equation}

\begin{equation}
P_{observation} = \frac{1}{\sqrt[]{2\pi\sigma^2}}e^\frac{-(x-\mu)^2}{2\sigma^2}
\label{eq:pdf}
\end{equation}

\begin{equation}
P_{emission,s,p} = P_{emission,p,i-1} * P_{transition,p,s} * P_{observation,s} 
\label{eq:emission}
\end{equation}

When calculating the probability of a fixation given a saccade input, the PDF function used to calculate the observation probability would occasionally return zero due to rounding errors. Being that the Viterbi algorithm builds its probabilities off of the previous probability, and that any number multiplied by zero returns zero, this can lead to poor results. Under this circumstance we reset the probability to an arbitrary lower bound of one ten-thousandth of the other state. In an instance that both probabilities equaled zero, a flag was set for the reset of both probabilities to an initial state. In a similar case, the PDF function would return a very small probability. After several iterations this would lead to a cascading effect where one classification would much outweigh the other using the same multiplicative probability property described above. In the instance the lower bound on probabilities has been exceeded, a flag was set to multiply both numbers by the log base 10 of the maximal probability, thus maintaining the probabilistic ratio between either state.

\begin{algorithm} 
     \SetAlgoLined
     \SetKwInOut{Input}{Input}\SetKwInOut{Output}{Output}
     \Input{array of eye position points, velocity threshold - $V_t$, dispersion threshold - $D_t$, minimum temporal window size - $W_t$}
     \Output{list of fixations, saccades, and smooth pursuits}
     Calculate velocities for each point\;
     Mark all points above $V_t$ as saccades\;
     Viterbi Algorithm(array of eye positions, [fixations, saccades], velocity, $V_t$)\;
      Filter saccades\;
      Initialize temporal window of size $W_t$ over remaining eye movements\;
      \While{temporal window not reaching end of array} {
      Calculate dispersion of points in window\;
        \eIf{dispersion < $D_t$} {
          \While{dispersion < $D_t$} {
          Mark point as fixation\;
          Add unclassified point to window\;
          Calculate dispersion of points in window\;
          }
        }{
      Mark first point as smooth pursuit\; 
      Move window to next point\;}
	}
    Viterbi Algorithm(array of eye positions, [fixations, smooth pursuits], dispersion, [$D_t$, $W_t$])\;
    Merge saccades back into eye position data\;
    Return list of classifications\;

\caption{I-VDT-HMM}
\label{alg:I-VDT-HMM}
\end{algorithm}

\begin{algorithm} 
     \SetAlgoLined
     \SetKwInOut{Input}{Input}\SetKwInOut{Output}{Output}
     \Input{array of eye position points, classes, feature, thresholds}
     \Output{list of classified classes}
     Calculate mean of feature for classes\;
     Calculate standard deviation of feature for classes\;
     Count transitions between each state\;
     \While{not converged}{
      Initialize emission and classification matrices\;
      Calculate observation probabilities\;
      Insert observation probabilities into first column of emission matrix\;
        \For{column in emission matrix > 1}{
          Calculate observation probabilities\;
          \If{observation probability == 0}{
          Set observation probability to lower bound\;
          }
          Calculate transitional probabilities\;
          Set classification matrix state to highest probability\;
          \If{emission probability < lower bound}{
          Normalize probabilities\;}
          
        }
        Calculate maximum final probability\;
        \For{last record to first record} {
          Set previous records highest probability as classification for current record\;
        }
        Calculate means and standard deviations of features\;
        Count state transitions\;
        \If{Previous iterations epsilon values are met}{
        converged\;
        }
       }
\caption{Viterbi Algorithm}
\label{alg:viterbi}
\end{algorithm}

\section{Methodology}
\subsection{Data}
The data used in our experiments was recorded using an EyeLink 1000 eyetracker recorded at 1000 Hz \cite{EyeLink1000} on a 21-in CRT monitor with a refresh rate of 80 Hz and a resolution of 1,024 by 768 pixels. The data consists of 11 subjects whom were recorded in monocular mode and produced various amounts of noise. The stimulus presented to the subjects was a 2-D step-ramp stimulus where the recorded data was converted into degrees of visual angle. The subject's eye records were classified into clean and noisy data, and given a ground truth label. This data was originally recorded, classified, and published in \cite{Komogortsev2013}, where you can find additional information.

\subsubsection{Data Subsampling}
As a high quality eye tracker is not always available, the data was subsampled into 7 different sampling frequencies: 30, 50, 60, 100, 200, 300, and 500 Hz. The three algorithms evaluated in this paper are tested on 8 of the 11 subjects at each of the 8 sampling frequencies. 3 of the subjects were dropped due to an issue with our parameter estimation algorithm. Of the data used we will focus primarily on the extremes for analysis of the reported algorithms: low, 30 Hz, and high, 1000 Hz, frequencies for clean, subject 007, and noisy, subject 010, datasets.

\subsection{Behavioral Scores}
We use the behavioral scores proposed in \cite{komogortsev2010standardization} and \cite{Komogortsev2013} in order to evaluate our algorithm due to their effectiveness in providing meaningful classifications in an automated setting. The behavioral scores provide a better result than direct classification accuracy as they take into account multiple factors found within a healthy individual's HOS which would affect such a classification method. For example, when considering a pursuits quantitative score (PQnS), the latency between seeing the target move and the HOS's response, the pursuit latency, as well as the time it takes for the HOS to catch up to the object are considered. The assumption then is made that the behavior of eye movements is matched with those of a healthy person's HOS. The scores measured in this work are the fixation quantitative score (FQnS), the saccade quantitative score (SQnS), pursuit quantitative score (PQnS), misclassified fixation score (MisFix), fixation qualitative score (PQlS), and the pursuit qualitative scores for positional (PQlS\_P) and velocity (PQlS\_V) accuracy. The ideal scores for each of the behavioral scores can be found in Table \ref{tab:IdealScores}.

The FQnS gives the means of computing the amount of fixation points classified correctly as fixations. We use the derivation of the ideal FQnS found in \cite{Komogortsev2013}, which takes into account the effects of SP on fixation classification. Through our calculations we found 81.6 to be the ideal behavioral score which differs slightly from the 83.87 score found in \cite{Komogortsev2013}.

The SQnS is defined as the ratio between the detected number of saccades and the total number of saccades in the stimulus \cite{komogortsev2010standardization}. In \cite{Komogortsev2013} the SQnS was modified to account for the detection of SPs in the stimulus using a temporal window over the step stimulus. This method considers anticipatory and corrective saccades. Ideally, the SQnS would be 100 as it would indicate the algorithm was able to successfully identify all saccades within the stimulus window.

The intuitive idea behind the PQnS is the ratio of correctly classified SPs over the total amount of SP stimulus. The ideal PQnS takes into account the latency and resulting corrective saccade exhibited by the HOS when given a SP stimulus. The calculation for ideal PQnS, 52.04, may be found in \cite{Komogortsev2013}.

The MisFix score is calculated as the amount of fixation points that are classified as a smooth pursuits over the total amount of fixation stimulus \cite{Komogortsev2013}. The ideal score takes into consideration the fact that the termination phase of a SP happens after the SP into the following fixation stimulus. This makes the assumption that each SP is followed by a fixation. The calculation for the ideal score can be found in \cite{Komogortsev2013}, which comes out to 7.1.

Intuitively, the FQlS is measured as the euclidean distance between a fixation stimulus and the detected fixation centroid \cite{komogortsev2010standardization}. The score is normalized by the total amount of points compared. The ideal score for FQlS is 0, however, the score is unlikely to be achieved due to inaccuracy of eye trackers and normal behavior of the HOS \cite{komogortsev2010standardization}.

The pursuit qualitative scores, PQlS\_P and PQlS\_V, like the FQlS, are used to compare the proximity as well as the velocity of the detected SPs to the corresponding stimulus \cite{Komogortsev2013}. The scores are then normalized over the amount of points compared. Due to calibration errors, corrective saccadic behaviors, HOS latency, and classification errors, the ideal scores of 0 may not be achievable \cite{Komogortsev2013}.

\subsection{Parameter estimation}
A parameter estimation algorithm was used in order to detect the ideal velocity and dispersion thresholds for the proposed algorithm. Every 5 velocity thresholds between 70 and 150 were used to optimize saccade detection during the initial phase of our algorithm. Testing every 0.1\degree\ between  0.1\degree\ and 2.0\degree's were used to optimize our dispersion threshold for the later part of our algorithm. As the minimum pause time of the eye is 200 ms \cite{salthouse1980determinants}, and that the minimum amount of time for the HOS to pick up any information is 100 ms \cite{widdel1984operational}, the assumption was made that while there is some variability in the HOS, at least 150 ms must pass while fixated in order to detect anything. Thus, a duration threshold of 150 ms was employed. As PQlS\_P and PQlS\_V are generally quite high due to catch up saccadic behavior, a multiplicative weight of 10 was added on to the difference between the classification and the ideal scores for the remaining behavioral scores.

\section{Results}
\subsection{I-BDT}
\begin{figure}[h]
\caption{Original I-BDT Code implemented in our environment and ran on subject s-007 on original 1000 Hz data }
\label{fig:IBDT_s007_1000_0}
\includegraphics[width=\columnwidth,height=5cm,keepaspectratio]{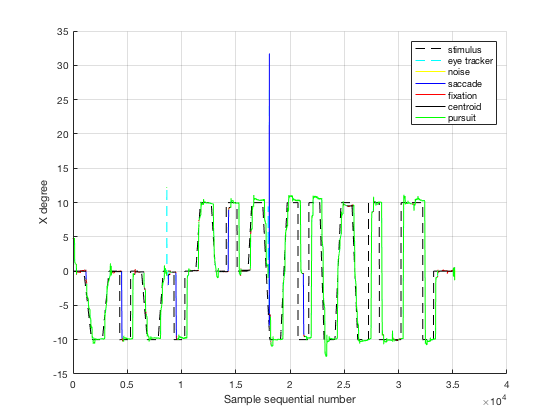}
\centering
\end{figure}

\begin{figure}[h]
\caption{Original I-BDT Code Implemented and Ran on Subject s-007 on Sampled 30 Hz Data }
\label{fig:IBDT_s007_30_0}
\includegraphics[width=\columnwidth,height=5cm,keepaspectratio]{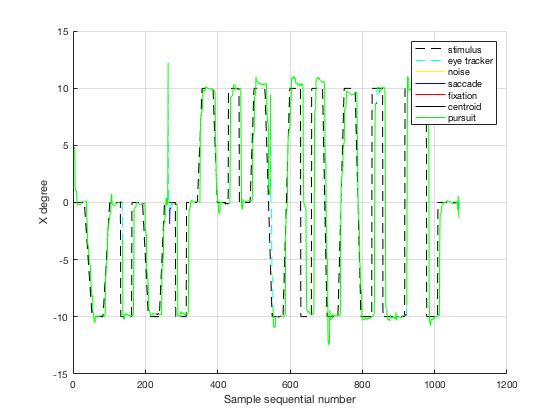}
\centering
\end{figure}

\begin{figure}[h]
\caption{I-BDT Results Using Multiple Fixation Thresholds on 1000 Hz Data}
\label{fig:IBDT_s007_1000_Mult_Fixation}
\includegraphics[width=\columnwidth]{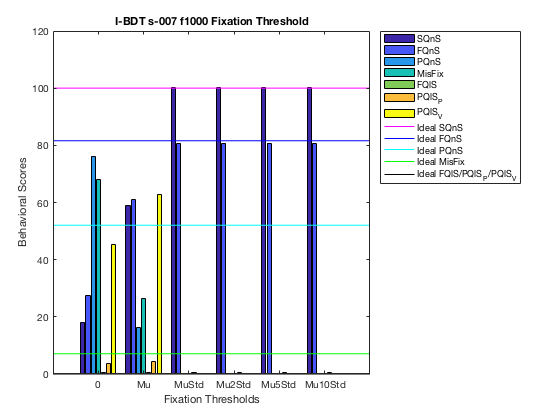}
\centering
\end{figure}

\begin{figure}[h]
\caption{I-BDT Results Using Multiple Fixation Thresholds on 30 Hz Data}
\label{fig:IBDT_s007_30_Mult_Fixation}
\includegraphics[width=\columnwidth]{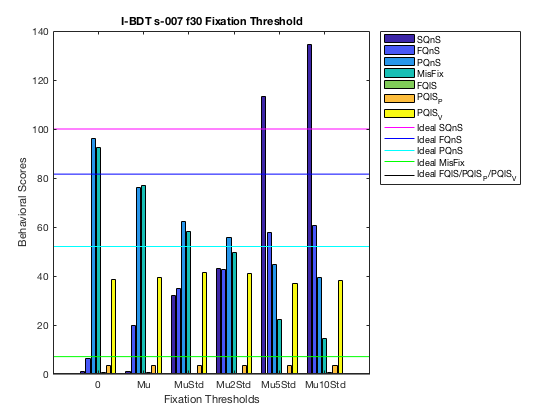}
\centering
\end{figure}

\begin{figure}[h]
\caption{I-BDT Results Using $\mu$ as Fixation Threshold on 1000 Hz Data}
\label{fig:IBDT_s007_1000_Mu}
\includegraphics[width=\columnwidth,height=5cm,keepaspectratio]{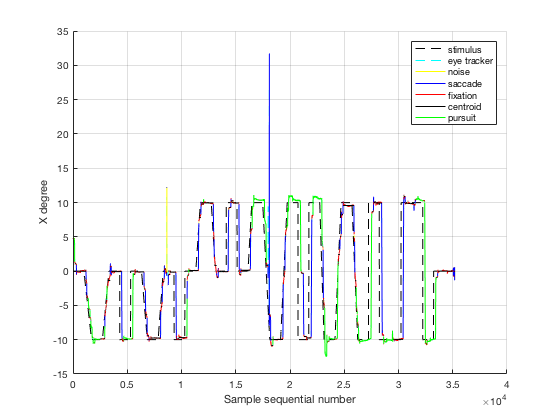}
\end{figure}

\begin{figure}[h]
\caption{I-BDT Results Using $\mu$ as Fixation Threshold Shown in Behavioral Scores on Clean and Noisy 30 Hz, 100 Hz, 500 Hz and 1000 Hz Data}
\label{fig:IBDT-optimal-results}
\includegraphics[width=\columnwidth]{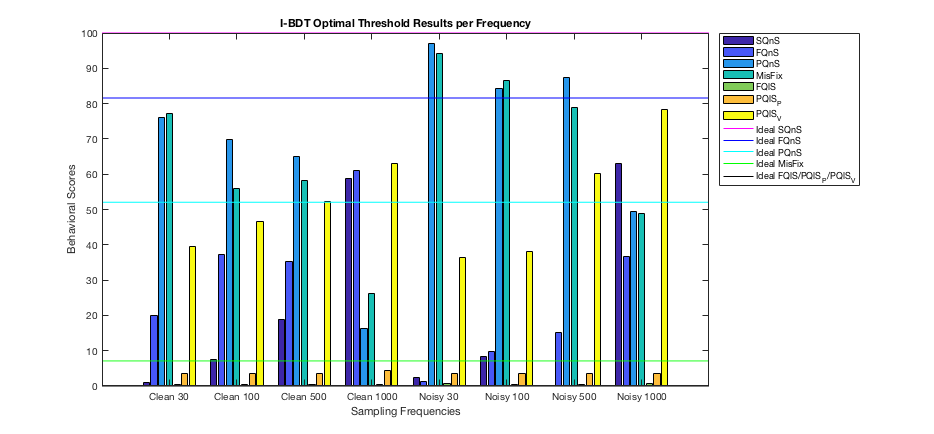}
\centering
\end{figure}

I-BDT was implemented in our environment on our data using the original code linked in \cite{Santini2016}. The classification results on clean 1000 Hz and 30 Hz data are shown in Figures \ref{fig:IBDT_s007_1000_0} and \ref{fig:IBDT_s007_30_0} respectively. The assumption made in \cite{Santini2016} was that fixations had a velocity of 0, and that the temporal window would be able to correctly discern between smooth pursuits and fixations. As eye trackers and the HOS have inherent noise associated with them, we tested the same 1000 Hz and 30 Hz data on a series of different fixation thresholds. The results are shown in Figures \ref{fig:IBDT_s007_1000_Mult_Fixation} and \ref{fig:IBDT_s007_30_Mult_Fixation} respectively. These results indicated that the fixation mean, $\mu$, may prove to provide better results on an average case as the fixation threshold when considering sample rates. The results of a fixation threshold of $\mu$ is shown on clean 1000 Hz data in Figure \ref{fig:IBDT_s007_1000_Mu}. Using $\mu$ as the fixation threshold, we were able to achieve the following results on clean and noisy data on sampling frequencies 30 Hz, 100 Hz, 500 Hz, and 1000 Hz shown in Figure \ref{fig:IBDT-optimal-results}.

\subsection{I-VDT}
\begin{figure}[h]
\caption{I-VDT Results on Subject 007 Sampled at 1000 Hz}
\label{fig:IVDT-s007-1000}
\includegraphics[width=\columnwidth,height=5cm,keepaspectratio]{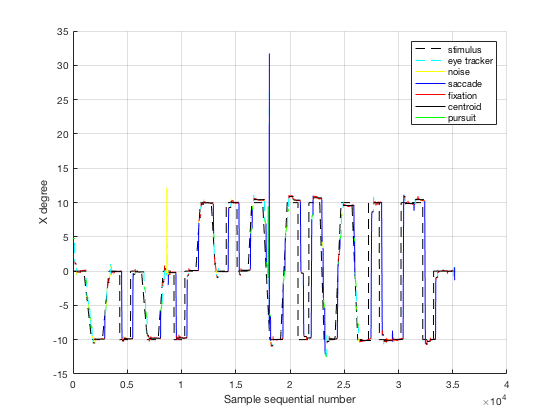}
\centering
\end{figure}

\begin{figure}[h]
\caption{I-VDT Results Using Optimal Thresholds Shown in Behavioral Scores on Clean and Noisy 30 Hz, 100 Hz, 500 Hz and 1000 Hz Data}
\label{fig:IVDT-optimal-results}
\includegraphics[width=\columnwidth]{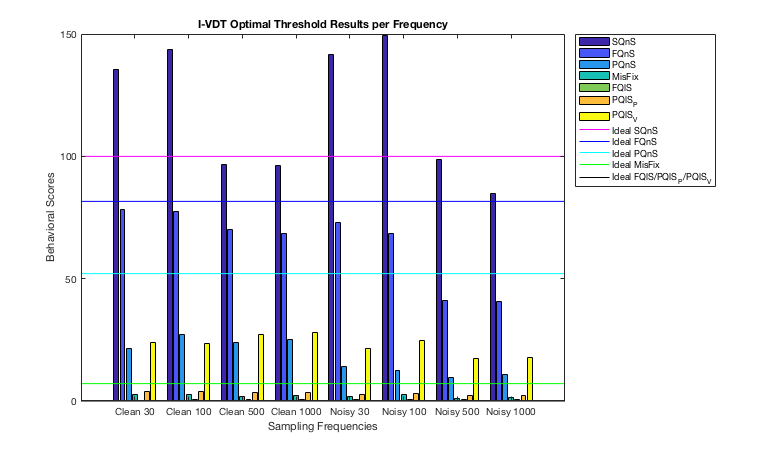}
\centering
\end{figure}

I-VDT was first proposed using the same environment and data we are using \cite{Komogortsev2013}. One difference is that we have subsampled the data to get an idea of how each algorithm is affected by sampling frequencies. The results of I-VDT on a clean dataset sampled at 1000 Hz is shown in Figure \ref{fig:IVDT-s007-1000}. Using the optimal thresholds detailed in \cite{Komogortsev2013}, we compiled the results on sampling frequencies of 30 Hz, 100 Hz, 500 Hz, and 1000 Hz on both clean and noisy data shown in Figure \ref{fig:IVDT-optimal-results}.

\subsection{I-VDT-HMM}
\begin{figure}[h]
\caption{I-VDT-HMM Optimal Thresholds Across Frequencies, Averaged Over Subjects}
\label{fig:IVDTHMM-optimal-thresholds}
\includegraphics[width=\columnwidth,height=5cm,keepaspectratio]{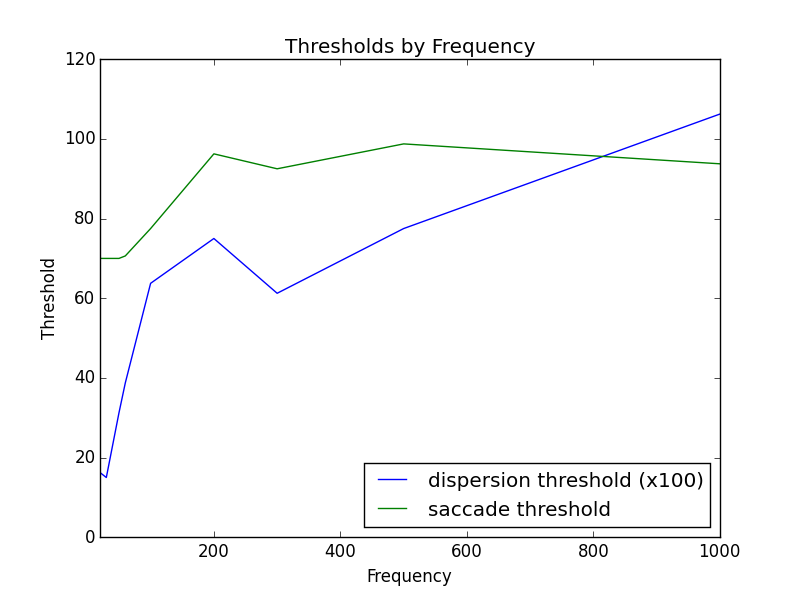}
\centering
\end{figure}

\begin{figure}[h]
\caption{I-VDT-HMM Optimal Threshold Results given in Behavioral Scores on Clean and Noisy 30 Hz, 100 Hz, 500 Hz, and 1000 Hz Data}
\label{fig:IVDTHMM-optimal-threshold-results}
\includegraphics[width=\columnwidth]{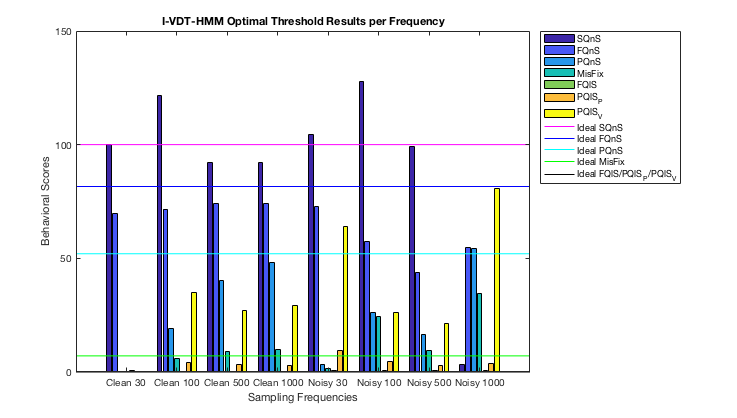}
\centering
\end{figure}

\begin{figure}[h]
\caption{I-VDT-HMM Optimal Threshold Classification Results on Subject 007 at 1000 Hz}
\label{fig:IVDTHMM-optimal-threshold-classification-results-1000}
\includegraphics[width=\columnwidth,height=5cm,keepaspectratio]{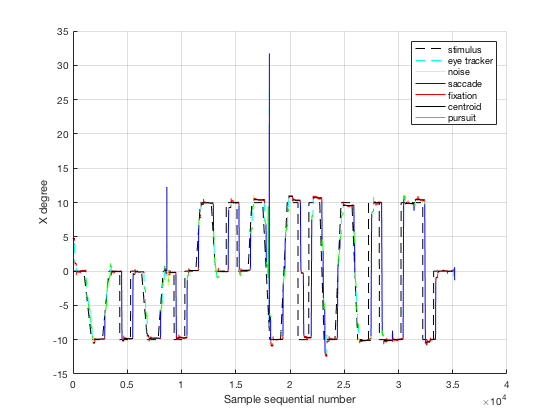}
\centering
\end{figure}

\begin{figure}[h]
\caption{I-VDT-HMM Optimal Threshold Classification Results on Subject 007 at 30 Hz}
\label{fig:IVDTHMM-optimal-threshold-classification-results-30}
\includegraphics[width=\columnwidth,height=5cm,keepaspectratio]{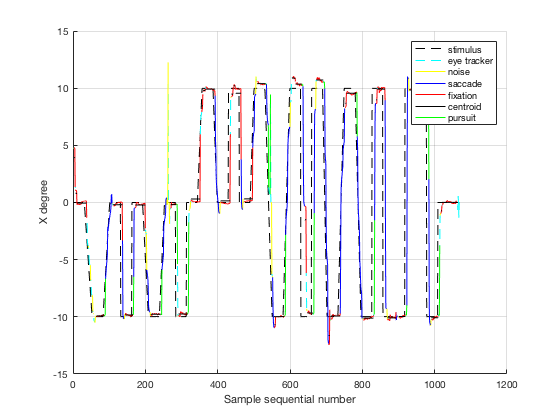}
\centering
\end{figure}

The proposed algorithm in this work, I-VDT-HMM, was tested on all of our subjects and frequencies in order to determine its optimal thresholds. The results which are averaged across the subjects are shown in Figure \ref{fig:IVDTHMM-optimal-thresholds}. The optimal thresholds selected for our data are shown in Table \ref{tab:optimalThresholds}. These optimal thresholds give the behavioral scores as shown in Figure \ref{fig:IVDTHMM-optimal-threshold-results} on subjects 007 and 010 across frequencies 30 Hz, 100 Hz, 500 Hz, and 1000 Hz. When using these thresholds on eye record data, we achieve the classification results using clean, subject 007, data on 1000 Hz and 30 Hz sampling frequencies shown in Figures \ref{fig:IVDTHMM-optimal-threshold-classification-results-1000} and \ref{fig:IVDTHMM-optimal-threshold-classification-results-30} respectively. The behavioral scores attributed to these classification results can be found in Table \ref{tab:behavioralScoresIVDTHMM}.

\begin{table}
\begin{tabular}{ |c|c| }
\hline
 Behavioral Score & Ideal Score \\
 \hline
 FQnS & 81.6\\
 SQnS & 100\\
 PQnS & 52.04\\ 
 MisFix & 7.1\\
 PQlS & 0\\
 PQlS\_P & 0\\ 
 PQlS\_V & 0\\
 \hline
\end{tabular}
\centering
\caption{Ideal Behavioral Scores}
\label{tab:IdealScores}
\end{table}

\begin{table}
\begin{tabular}{ |c|c| }
\hline
 Velocity Threshold & 75 \\
 Dispersion Threshold & 0.67 \\
 Duration Threshold & 150 \\ 
 \hline
\end{tabular}
\caption{Optimal Thresholds for I-VDT-HMM}
\label{tab:optimalThresholds}
\centering
\end{table}

\begin{table}
\begin{tabular}{ |c|c|c| }
\hline
Behavioral Score & 1000 Hz Result & 30 Hz Result\\
\hline
 FQnS & 73.91 & 69.41 \\
 SQnS & 92.13 & 102.06 \\
 PQnS & 48.26 & 12.43\\
 MisFix & 10.01 & 3.82\\
 FQlS & 0.39 & 0.43\\
 PQlS\_P & 3.06 & 6.02\\
 PQlS\_V & 29.34 & 86.00\\
\hline
\end{tabular}
\caption{Behavioral Scores using I-VDT-HMM for Subject 007 on 1000 Hz and 30 Hz Data}
\label{tab:behavioralScoresIVDTHMM}
\centering
\end{table}

\subsection{Cross Algorithm Results}
\begin{figure}[h]
\caption{Comparison of Algorithms Mean Classification Times by Frequency}
\centering
\label{fig:classificationTime}
\includegraphics[width=\columnwidth,height=5cm,keepaspectratio]{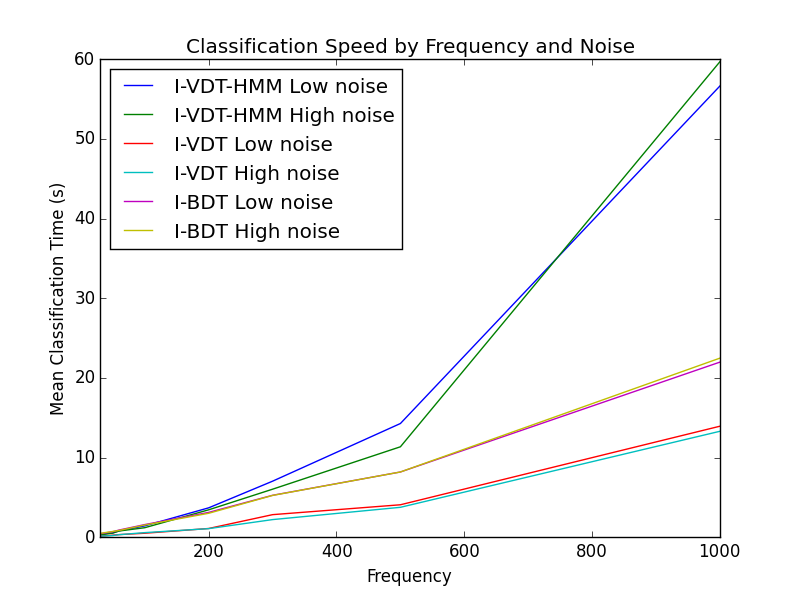}
\end{figure}

\begin{figure}[h]
\caption{Comparison of Algorithms on Subject 007 Data using 1000 Hz Sample Frequency}
\label{fig:algorithms-007-1000}
\includegraphics[width=\columnwidth]{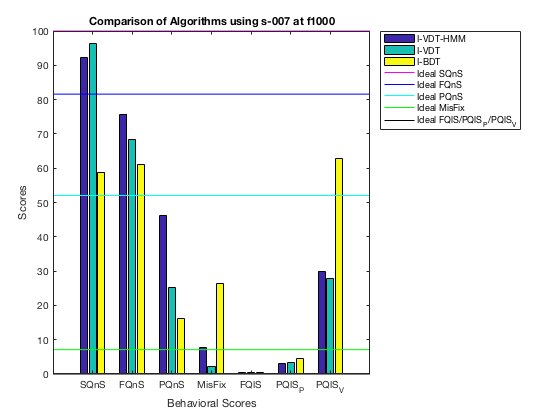}
\end{figure}

\begin{figure}[h]
\caption{Comparison of Algorithms on Subject 007 Data using 30 Hz Sample Frequency}
\centering
\label{fig:algorithms-007-30}
\includegraphics[width=\columnwidth]{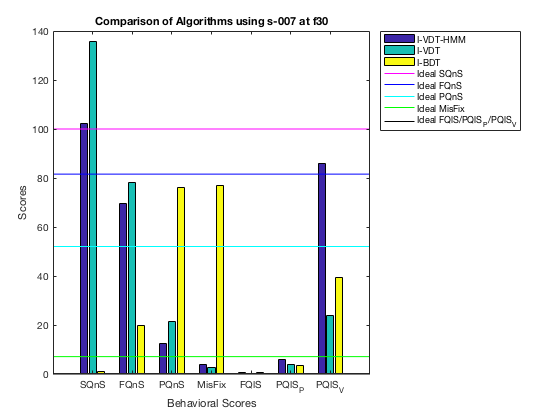}
\end{figure}

\begin{figure}[h]
\caption{Comparison of Algorithms on Subject 010 Data using 1000 Hz Sample Frequency}
\centering
\label{fig:algorithms-010-1000}
\includegraphics[width=\columnwidth]{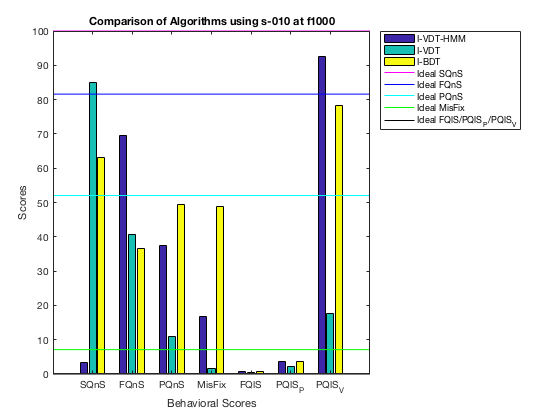}
\end{figure}

\begin{figure}[h]
\caption{Comparison of Algorithms on Subject 010 Data using 30 Hz Sample Frequency}
\centering
\label{fig:algorithms-010-30}
\includegraphics[width=\columnwidth, keepaspectratio]{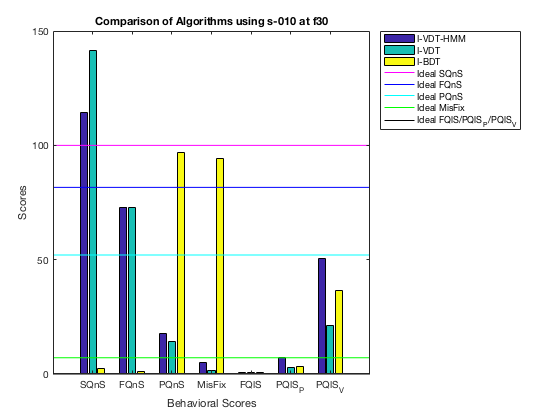}
\end{figure}

Figure \ref{fig:classificationTime} presents the average classification time between algorithms over all tested frequencies and subjects. The X axis represents the frequency the eye records were sampled to, and the Y axis represents the classification time. Figures \ref{fig:algorithms-007-1000} and \ref{fig:algorithms-007-30} compare all algorithms using the clean data set, provided by subject 007, while Figures \ref{fig:algorithms-010-1000} and \ref{fig:algorithms-010-30} compare all algorithms using the noisy data set, provided by subject 010, on sample frequencies 1000 Hz and 30 Hz respectively.

\section{Discussion}
On a clean dataset, as shown in the original paper \cite{Santini2016}, I-BDT was able to achieve outstanding scores averaging 94.98\% classification accuracy. On the basis of Figures \ref{fig:IBDT_s007_1000_0} and \ref{fig:IBDT_s007_30_0}, we can determine that the assumption made about fixation velocities being equal to zero to be poorly made when introduced to a relatively noisy dataset. When determining a more optimal fixation threshold, we tested fixation thresholds 0, $\mu$, and several $\mu$'s plus multiples of the standard deviation to see how the scores would fare. For lower sampling rates, the higher the fixation threshold the better the algorithm did. This is likely a result of how noisy data can affect the velocity of points. For higher sampling rates after averaging the overall thresholds we found that the mean velocity of a fixation proved to be the most reliable. Using the mean velocity for the fixation threshold we were able to produce the classification results for clean 1000 Hz data found in Figure \ref{fig:IBDT_s007_1000_Mu}.

Through the algorithm comparison figures, \ref{fig:algorithms-007-1000}, \ref{fig:algorithms-007-30}, \ref{fig:algorithms-010-1000} and \ref{fig:algorithms-010-30}, we can see that I-VDT and I-VDT-HMM are the most closely matched algorithms, with I-VDT-HMM obtaining a closer-to-ideal score roughly half of the time. As expected, in Figure \ref{fig:classificationTime} we can see that I-VDT is substantially faster than I-VDT-HMM. I-VDT-HMM aims to optimize results at the cost of time.

Observing Figure \ref{fig:algorithms-010-1000} we can see that the SQnS score is quite low on the high frequency noisy data for I-VDT-HMM. An issue with the Viterbi Algorithm \cite {forney1973viterbi} convergence method is that when introduced to data that has too much noise, the noise will greatly affect the mean velocity of each of the states. Being that the mean velocity is included in our epsilon values, adding additional data points to a class will affect its mean. Assuming the fixation velocity starts out high, as to be expected in noisy data, the mean velocity of saccades will continue to rise until the only saccades left are near peak velocity. Once the mean becomes too high, the PDF function will eventually return a 0 result for fixations leading to our aforementioned PDF issue. It is interesting to note that when this data is subsampled, the scores become much better as can be seen in Figure \ref{fig:algorithms-010-30}. This is likely due to much of the noise being filtered out, leaving a more clean eye record to classify.

\section{Conclusion}
In this work we provide an analysis of multiple tertiary eye movement classification algorithms: I-VDT \cite{Komogortsev2013}, a threshold based algorithm, I-BDT \cite{Santini2016}, a probability based algorithm, and a newly proposed algorithm: I-VDT-HMM, a hybrid threshold and probability based algorithm. We use the behavioral scores introduced in \cite{komogortsev2010standardization} and \cite{Komogortsev2013} to assess the quality of each algorithm due to their accounting of normal behavior associated with the HOS while also providing automated threshold selection. Our findings show that the proposed algorithm, I-VDT-HMM, has promising results on high frequency low noise data while performing poorly on noisy data.

The next thing that will be done is deriving a solution to the round-to-zero error we're experiencing with the PDF function in calculating the observational probabilities as it will vastly improve the algorithm's ability to handle noise. In the future, we will test I-VDT-HMM on more subjects and use cross-fold validation in order to acquire more statistically significant results. Using precision recall and F1 scores would provide a secondary method of algorithm comparison. It would be interesting to see the results that would come from the forward-backward algorithm applied in a similar way, as well as compare the results of I-HMM \cite{Salvucci2000}.

\clearpage

\bibliographystyle{ACM-Reference-Format}
\bibliography{sample-bibliography} 

\end{document}